\documentclass[runningheads]{llncs}

\usepackage[T1]{fontenc}
\usepackage{color}
\usepackage{amsmath}
\usepackage{amsfonts}
\usepackage{amssymb}
\usepackage{float}
\usepackage{xcolor}
\usepackage{subfig}
\usepackage{graphicx}
\usepackage{siunitx}

\usepackage{tikz}
\usepackage[paperheight=11in,paperwidth=8.5in,margin=1in]{geometry}
\usetikzlibrary{calc,arrows.meta,tikzmark}

\newcommand{\hlbox}[2]{\colorbox{#1!25}{\strut #2}}
\newcommand{\markhl}[3][green]{\tikzmarknode{#2}{\hlbox{#1}{#3}}}

\newcommand{\linkhlsoft}[3][gray]{%
  \begin{tikzpicture}[remember picture,overlay]
    \draw[#1!80, line width=0.2pt]
      (#2) to[bend left=8] (#3);
  \end{tikzpicture}%
}

\usepackage{hyperref}
\hypersetup{
    colorlinks=true,
    linkcolor=blue,
    filecolor=blue,      
    urlcolor=blue,
    citecolor=blue,
}

\begin{document}

\title{Distance-to-Distance Ratio: A Similarity Measure for Sentences Based on Rate of Change in LLM Embeddings}

\titlerunning{Distance-to-Distance Ratio}

\author{Abdullah Qureshi\inst{1} \and
	Kenneth Rice\inst{1} \and
	Alexander Wolpert\inst{2}
	}

\authorrunning{A. Qureshi et al.}

\institute{DataKnife, Chicago IL 60611, USA\\
	\email{\{abdullah,kenny\}@dataknife.io}\\
	\url{https://www.dataknife.io} \and
	Roosevelt University, Chicago IL 60605, USA\\
	\email{awolpert@roosevelt.edu}\\
	\url{https://www.roosevelt.edu}}

\maketitle
\begin{abstract}
	A measure of similarity between text embeddings can be considered adequate only if it adheres to the human perception of similarity between texts. In this paper, we introduce the distance-to-distance ratio (DDR), a novel measure of similarity between LLM sentence embeddings. Inspired by Lipschitz continuity, DDR measures the rate of change in similarity between the pre-context word embeddings and the similarity between post-context LLM embeddings, thus measuring the semantic influence of context. We evaluate the performance of DDR in experiments designed as a series of perturbations applied to sentences drawn from a sentence dataset. For each sentence, we generate variants by replacing one, two, or three words with either synonyms, which constitute semantically similar text, or randomly chosen words, which constitute semantically dissimilar text. We compare the performance of DDR with other prevailing similarity metrics and demonstrate that DDR consistently provides finer discrimination between semantically similar and dissimilar texts, even under minimal, controlled edits.

\end{abstract}

\section{Introduction}
\label{sec:Intro}
For two pieces of text to be considered semantically similar they must generally "mean the same thing," irrespective of whether they use precisely the same words. \textit{The cat enjoyed chasing the mouse} is semantically similar to \textit{the feline liked running after the rodent} and it is semantically dissimilar to \textit{it is raining out.}

In current applications, domain objects are generally represented by object embeddings within a mathematical structure equipped with a similarity or distance function. Therefore, for an embedding to be considered truly semantic, the (dis)similarity between objects should correspond to (dis)similarity between embeddings. It follows then that for an embedding to truly encode semantic meaning, it should also respect such notions of semantic continuity: small semantic edits should produce similarly small shifts in the geometry of embedded objects, while large semantic changes should result in commensurately large geometric shifts.

Most LLM pipelines map each text to a single vector and compare vectors with cosine distance. Two standard representations include centroid (or mean pooling), which uses the mean of all token embeddings and is widely utilized for summarization and document similarity\cite{lamsiyahUnsupervisedMethodExtractive2021,pmlr-v37-kusnerb15}; and EOS, in which a final end‑of‑sequence token vector accumulates contextual information from all previous tokens, commonly utilized in decoder‑only models\cite{leeNVEmbedImprovedTechniques2025}. Despite its popularity, the reliability of cosine similarity in measuring semantic similarity from contextual embeddings alone remains inconclusive\cite{steckCosineSimilarityEmbeddingsReally2024,youSemanticsAngleWhen2025,zhouProblemsCosineMeasure2022}. Due to this uncertainty, we propose a method that not only serves as a more sensitive evaluator of semantic shifts, but also evaluates how effectively different embedding techniques and distance measures capture semantic similarity.

In this paper, we introduce the distance‑to‑distance ratio (DDR), which tracks how token‑pair distances change from pre- to post‑context embeddings, drawing on the concept of Lipschitz continuity. Following the recent paper\cite{pmlr-v54-kpotufe17a}, ratios can be substantially easier to estimate than the underlying quantities themselves. In particular, density-ratio estimation depends only on the smoothness of the ratio (which can be far greater than that of the individual densities) and can even achieve parametric rates in favorable cases\cite{pmlr-v54-kpotufe17a}. This motivates our choice in using the distance-to-distance ratio, which is explained below. Rather than using pre- and post-context distances separately, we directly compute their ratio, which is theoretically more stable in high-dimensional settings. DDR's effectiveness depends on the ratio’s smoothness and the data’s intrinsic dimension, not the often large ambient dimension. Consequently, DDR detects subtle semantic changes more robustly than cosine distance on context embeddings alone.

Structure. Section~\ref{sec:prelim} reviews embeddings and metrics; Section~\ref{sec:DDR} formalizes DDR; Section~\ref{sec:design} details the experimental setup; Section~\ref{sec:results} presents results; Section~\ref{sec:conclusion} outlines extensions and future applications.

\section{Preliminaries}
\label{sec:prelim}

\subsection{Distances}
\label{subsec:distances}

\subsubsection{Cosine}
\label{subsubsec:cosine}
Given two input vectors \(\boldsymbol{\mu}_X\) and \(\boldsymbol{\mu}_{X'}\), the cosine similarity between them can be computed thus:
\begin{equation}
    \label{eq:cos_def}
    \textstyle \text{s}_{\cos}(\boldsymbol{\mu}_X, \boldsymbol{\mu}_{X'}) = \frac{\boldsymbol{\mu}_X \cdot \boldsymbol{\mu}_{X'}}{\|\boldsymbol{\mu}_X\|\|\boldsymbol{\mu}_{X'}\|}.
\end{equation}

Whenever it makes sense for comparison purposes, we also use cosine distance (a complement of cosine similarity):
\begin{equation}
    \text{d}_{\cos}(\boldsymbol{\mu}_X, \boldsymbol{\mu}_{X'}) = 1 - \text{s}_{\cos}(\boldsymbol{\mu}_X, \boldsymbol{\mu}_{X'}).
\end{equation}

Note that cosine distance is not a metric on $\mathbb{R}^n$.

\subsubsection{Chordal}
\label{subsubsec:Chordal}
Let \(\mathbf{u}, \mathbf{v} \in\mathbb{R}^d\) each be nonzero vectors. After normalizing both to the unit sphere, chordal distance is defined as the straight‐line (Euclidean) distance between their tips.

In terms of cosine similarity, it can be written as:
\begin{equation}
    \text{d}_{\mathrm{chord}}(\mathbf{u}, \mathbf{v}) = \sqrt{2\Bigl(1 - \frac{\mathbf{u} \cdot \mathbf{v}}{\|\mathbf{u}\|\|\mathbf{v}\|}\Bigr)} = \sqrt{2\Bigl(1 - \text{s}_{\cos}(\mathbf{u}, \mathbf{v})\Bigr)}.
\end{equation}

Here, \( \text{s}_{\cos}(\mathbf{u}, \mathbf{v})\) is defined above~\ref{eq:cos_def}.

\subsubsection{Earth Mover's Distance}
\label{subsubsec:EMD}
In this formulation, consider two discrete experimental signature distributions\cite{rubnerMetricDistributionsApplications1998}:
\begin{equation}
\begin{aligned}
    P &= \{(\mathbf{p}_{1}, w_{p_1}), (\mathbf{p}_{2}, w_{p_2}), \dots, (\mathbf{p}_{m}, w_{p_m})\},\\[6pt]
    Q &= \{(\mathbf{q}_{1}, w_{q_1}), (\mathbf{q}_{2}, w_{q_2}), \dots, (\mathbf{q}_{n}, w_{q_n})\}.
\end{aligned}
\end{equation}

Let \(D = [\,d_{i,j}\,]\) be the ground distance between \(\mathbf{p}_i\) and \(\mathbf{q}_j\). Then the Earth Mover’s Distance between \(P\) and \(Q\) is calculated based on the optimal flow \(F=[f_{i,j}]\), with \(f_{i,j}\) representing the flow between \(\mathbf{p}_i\) and \(\mathbf{q}_j\), that minimizes the overall cost:
\begin{equation} 
    \min_{F}\sum_{i=1}^{m}\sum_{j=1}^{n}f_{i,j}\,d_{i,j}
\end{equation}
subject to the constraints:
\begin{equation}
    \begin{aligned}
        & f_{i,j} \ge 0,                        && 1\le i\le m,\; 1\le j\le n, \\
        & \sum_{j=1}^{n} f_{i,j} \leq w_{p_i},  && 1\leq i\leq m,              \\
        & \sum_{i=1}^{m} f_{i,j} \leq w_{q_j},  && 1\leq j\leq n,              \\
        & \sum_{i=1}^{m}\sum_{j=1}^{n} f_{i,j} = \min\Bigl\{ \sum_{i=1}^m w_{p_i},\; \sum_{j=1}^n w_{q_j} \Bigr\}.
    \end{aligned}
\end{equation}

The optimal flow \(F\) is found by solving this linear optimization problem. The Earth Mover's Distance is defined as the work normalized by the total flow:

\begin{equation}
    \text{EMD}(P,Q)
    = \frac{\sum_{i=1}^{m}\sum_{j=1}^{n} f_{i,j}\,d_{i,j}}
    {\sum_{i=1}^{m}\sum_{j=1}^{n} f_{i,j}}
\end{equation}

\subsection{Embedding Pooling Methods}
\label{subsec:Embeddings}

\subsubsection{Centroid}
\label{subsubsec:centroid}
The Centroid (or mean pooling) method summarizes each sequence into one high-dimensional vector by averaging the embeddings from the final hidden layer\cite{leeNVEmbedImprovedTechniques2025}.

Formally, to calculate this method, let \( X = \{\mathbf{x}_1, \mathbf{x}_2, \dots, \mathbf{x}_n\} \in \mathbb{R}^{n \times d} \) be the sequence of contextual embeddings for \(n\) tokens, where each \(\mathbf{x}_i \in \mathbb{R}^{d}\) is the embedding of the \(i\)-th token, and \(d\) is the embedding dimension.

We then define the centroid vector \(\boldsymbol{\mu}_X \in \mathbb{R}^d\) as:
\begin{equation}
    \boldsymbol{\mu}_X = \frac{1}{n}\sum_{i=1}^{n}\mathbf{x}_i.
\end{equation}
\subsubsection{EOS}
The EOS method generates a single sequence embedding using the last hidden-layer \texttt{<EOS>} token vector, representing the cumulative meaning of the entire input\cite{wangImprovingTextEmbeddings2024}\cite{leeNVEmbedImprovedTechniques2025}.

Given a sequence of contextual embeddings
\begin{equation}
    X = \{\mathbf{x}_1, \mathbf{x}_2, \dots, \mathbf{x}_n, \mathbf{x}_{\text{eos}}\} \in \mathbb{R}^{(n+1)\times d}
\end{equation}
we select the EOS embedding as \(\mathbf{x}_{\text{eos}} \in \mathbb{R}^{d}\).

\section{Distance-to-Distance Ratio}
\label{sec:DDR}
As we show empirically in Section~\ref{sec:results}, the common approaches of cosine similarity between contextual embeddings generally fail to discriminate between (dis)similar texts. Given the hypothesis that similar tokens should be closer together and dissimilar tokens should remain apart, we conclude that these approaches are not suitable for this purpose.

Therefore, inspired by the Lipschitz constant for functions and Lipschitz-style distance expansion, we propose a method that measures the rate at which (dis)similarity between initial embedding positions transforms into (dis)similarity between output vector positions. While the Lipschitz constant measures the rate of change induced by a function, Lipschitz-style distance expansion measures the rate of change in distance between vectors\cite{pmlr-v54-kpotufe17a}. With this in mind, DDR measures the in-to-out rate of change created by the LLM attention transformation. As stated in Section~\ref{sec:Intro}, it has been shown that ratios can often be estimated more robustly than the underlying quantities themselves\cite{pmlr-v54-kpotufe17a}; DDR leverages this principle to provide a more stable evaluation of semantic shift.

\textbf{Geometric Intuition.} We hypothesize that synonym substitutions move along the low-dimensional manifold of coherent text, where the model is locally contractive (yielding high DDR). Conversely, random substitutions likely force the representation off-manifold, causing divergent output changes and lower DDR values.

To calculate DDR, we compute the maximum (over token positions) of the distance between the two texts’ vectors at matching positions in the input embeddings, and divide it by the analogous maximum distance after the LLM transformation. In doing so, DDR measures how much the transformation contracts or dilates.

\subsection{Calculating Distance-to-Distance Ratio}

To formally introduce the distance-to-distance ratio (DDR), we define a product
metric on sequences of the same length. Fix $L\in\mathbb{N}$ and let
$x=(\mathbf{x}_1,\ldots,\mathbf{x}_L)$, $x'=(\mathbf{x}'_1,\ldots,\mathbf{x}'_L)$ with $\mathbf{x}_i\in\mathbb{R}^m$.
Define
\begin{equation}
    d(x,x') = \max_{1\le i\le L} \text{d}_{\text{chord}}(\mathbf{x}_i,\mathbf{x}'_i).
\end{equation}
It is well known (see \cite{dezaEncyclopediaDistances2013}) that, \(\text{d}_{\text{chord}}\) is a metric on normalized vectors.

Let $F:\mathbb{R}^{m\times L}\to\mathbb{R}^{n\times L}$ be an LLM that maps
$x=(\mathbf{x}_1,\ldots,\mathbf{x}_L)$ to embeddings $y=(\mathbf{y}_1,\ldots,\mathbf{y}_L)=F(x)$,
with $\mathbf{y}_i\in\mathbb{R}^n$ (the parameters $m,n$ are fixed for $F$).
When comparing sequences, we use the same length $L$ for the inputs $x,x'$ and the embeddings $y,y'$.
Define
\begin{equation}
    d_{\text{in}}(x,x') = \max_{1\le i\le L} \text{d}_{\text{chord}}(\mathbf{x}_i,\mathbf{x}'_i),
    \qquad
    d_{\text{out}}(y,y') = \max_{1\le i\le L} \text{d}_{\text{chord}}(\mathbf{y}_i,\mathbf{y}'_i).
\end{equation}

Let $t,t'$ be the texts that a tokenizer transforms into embeddings $x, x'$ respectively. We then define DDR on texts $t,t'$ of the same length as
\begin{equation}
    \label{eq:DDR_def}
    \text{DDR}(t,t') = \frac{d_{\text{in}}(x,x')}{d_{\text{out}}(F(x),F(x'))},
\end{equation}
where $F$ is the LLM transformation, and $F(x),F(x')$ are the respective LLM embeddings.

\section{Experiment Design}
\label{sec:design}
We assess how well GPT‑style embeddings capture semantic similarity by progressively perturbing each of 500 source texts into three groups that have edits at three depths, with one, two, and three words substituted. For every perturbation we produce two modified texts to compare to the original, one with synonym substitutions and another with random word substitutions, where the number of words substituted is based on the edit depth. We then compute similarity scores with DDR, Centroid, and EOS methods, yielding two distributions at each of the three edit depths: one for synonym substitutions and one for random substitutions. For the Centroid and EOS methods, the similarity score is defined as the cosine similarity (Eq.~\ref{eq:cos_def}) between the respective vector representations of the context embeddings; for the DDR method, the score is the raw ratio calculated by (Eq.~\ref{eq:DDR_def}). If embeddings encode meaning, synonym scores should be consistently higher than random ones.

\subsection{Analysis Techniques}
\label{subsec:analysis_techniques}
We evaluate each method’s ability to distinguish between synonym and random word substitutions. For each method and edit depth, we plot the empirical CDFs of similarity scores for synonym and random edits. A steeper rise for random edits compared to synonym edits indicates better discrimination; greater separation between the CDFs implies stronger distinction, while overlapping CDFs suggest poor differentiation.

To quantify the separation, we compute the Earth Mover’s Distance (EMD), which measures the minimum “work” needed to transform one distribution into another. For EMD, each empirical score distribution is treated as unit mass, and we interpret EMD within each method’s native score scale (i.e., we don’t compare values across methods). The EMD quantifies how far the synonym distribution is shifted relative to the random distribution. Larger EMD values indicate greater distinction (see Section~\ref{subsubsec:EMD}).

To visualize the difference we plot scatter diagrams where each point corresponds to a sentence pair, and to quantify the difference we calculate the Pearson coefficient. The y-values are the similarity scores under synonym substitution and the x-values are the similarity scores under random substitution. A method that distinguishes well will have points predominately above the y = x line and a low Pearson score, indicating de-correlation between synonym and random substitution similarities.

\subsection{Datasets}
Our dataset consists of 500 source excerpts from \textit{The Hacker Crackdown}\cite{sterlingHackerCrackdown1994} and \textit{Dracula}\cite{stokerDracula2004}. For each excerpt, we generate three variant sets corresponding to one, two, or three word replacements (edit depths). Within each set, we create two substitution types: one using synonyms and one using random words, ensuring that original, synonym-substituted, and random-substituted excerpts all have identical token lengths.

\begin{figure}
    \centering
    \includegraphics[width=0.99\textwidth]{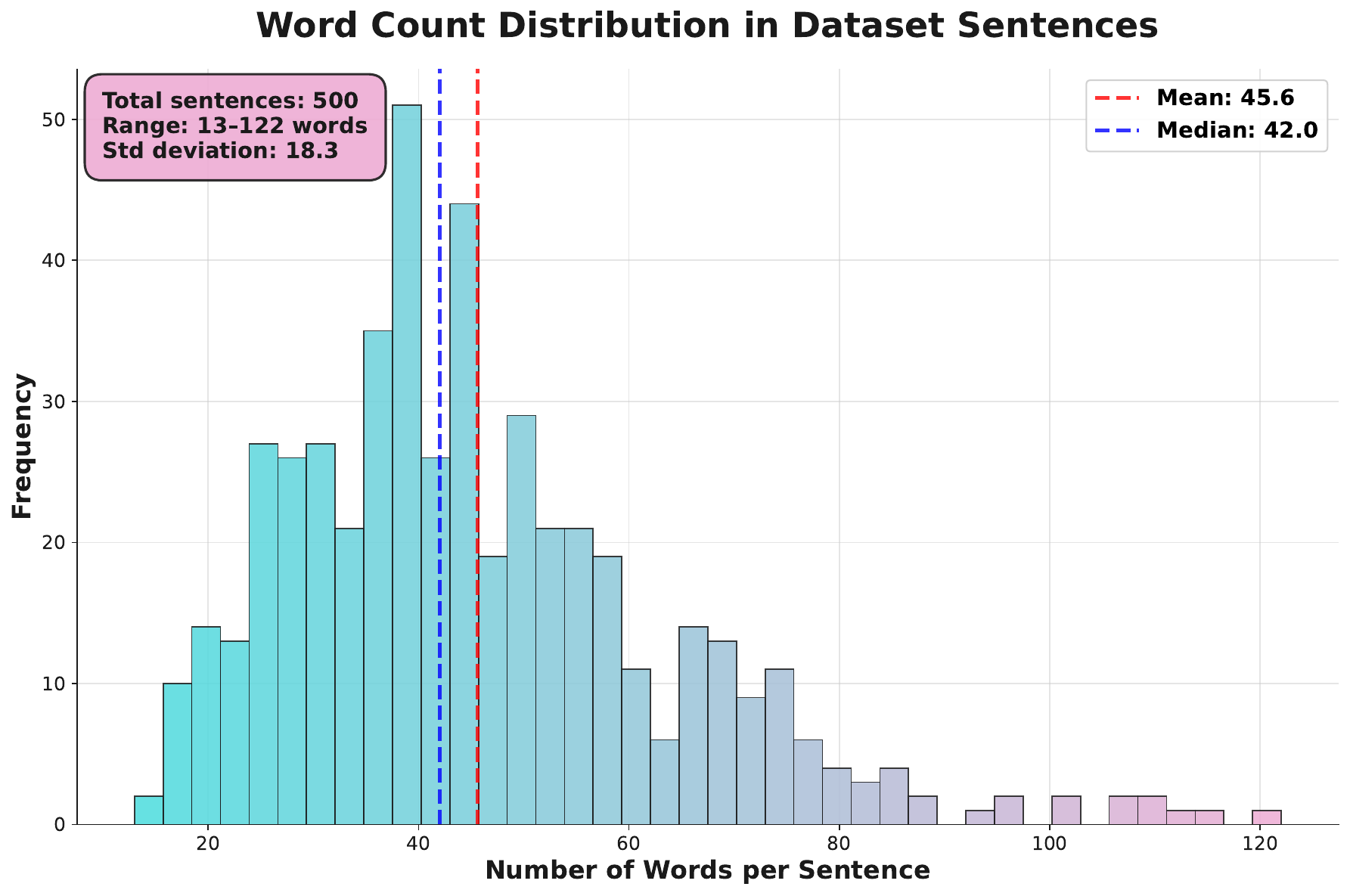} 
    \caption{Histogram of word counts}
    \label{fig:wordcount}
\end{figure}

Figure \ref{fig:wordcount} presents the distribution of word counts across the 500 source excerpts in our dataset. The distribution exhibits a roughly normal shape with a slight right skew, centered around a mean of 45.6 words per excerpt (median: 42 words). The majority of excerpts contain between 31 and 50 words, with relatively few outliers at the extremes (minimum: 13 words, maximum: 122 words, standard deviation: 18.3). This moderate length range provides sufficient context for meaningful semantic analysis while avoiding the potential confounding effects of extremely short fragments or overly long passages. The consistency in excerpt length helps ensure that our similarity measurements are not unduly influenced by length disparities between compared texts.

\subsubsection{Single word edit example}

{%
  \centering
  Police shudder at this prospect. Police treasure
  \markhl[teal]{A}{good} relations with the business community.\par
}\vspace{1em}

\noindent
\begin{minipage}[t]{0.48\linewidth}
\centering\textbf{Synonym Substitution}\\[0.5em]
Police shudder at this prospect. Police treasure \markhl[cyan]{A2}{right} relations with the business community.
\end{minipage}\hfill%
\begin{minipage}[t]{0.48\linewidth}
\centering\textbf{Random Substitution}\\[0.5em]
Police shudder at this prospect. Police treasure \markhl[red]{A3}{than} relations with the business community.
\end{minipage}

\linkhlsoft[cyan]{A}{A2}
\linkhlsoft[red]{A}{A3}

\vspace{1.5\baselineskip}

\par\noindent

\section{Results}
\label{sec:results}
\begin{figure}[!ht]
	\centering
	\begin{tabular}[c]{ccc}
		\subfloat[Centroid and Cosine]{\label{fig:1a}                                       
		\begin{tabular}[b]{c}                                                               
		\includegraphics[width=0.31\textwidth]{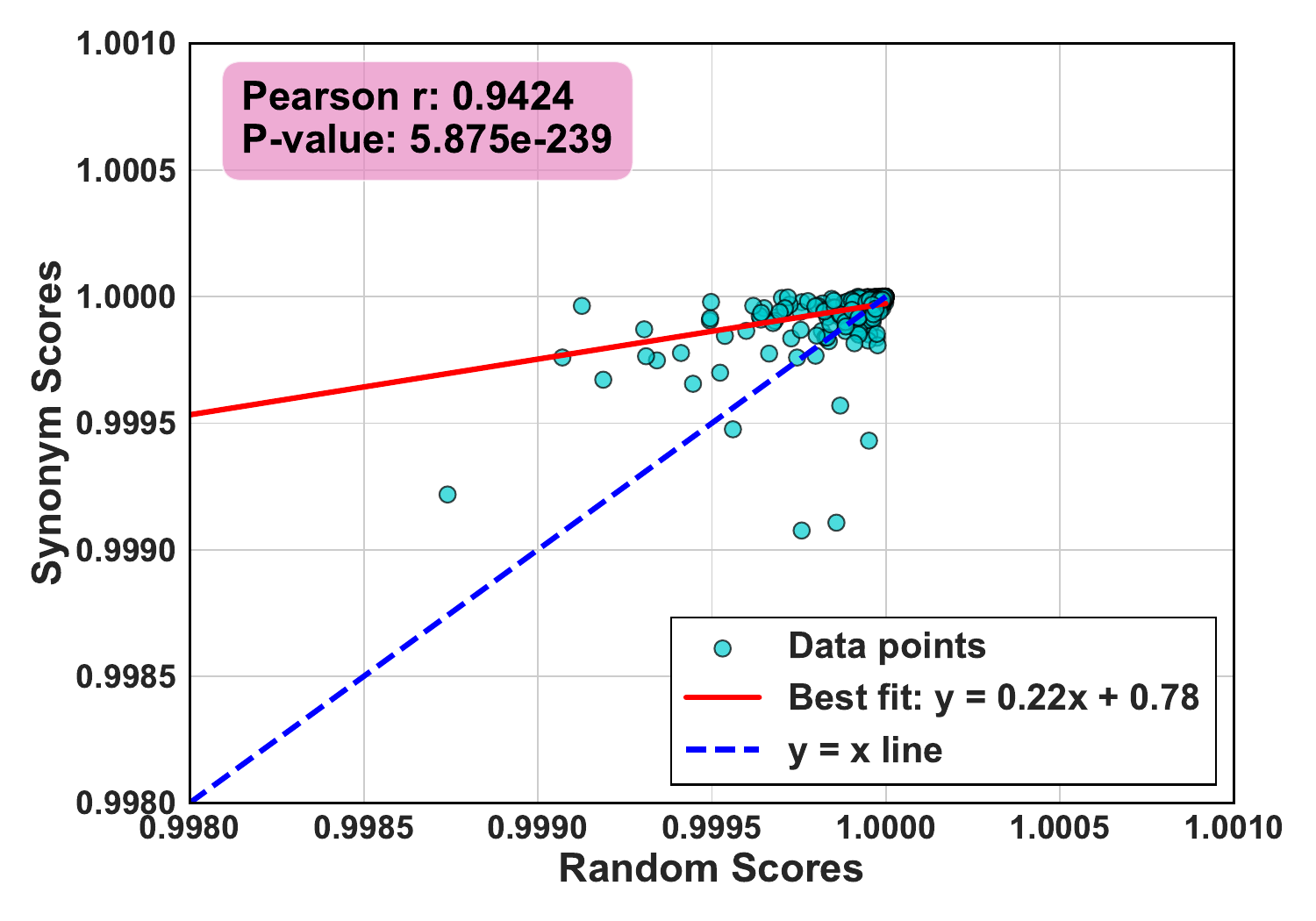} \\[2mm]
		\includegraphics[width=0.31\textwidth]{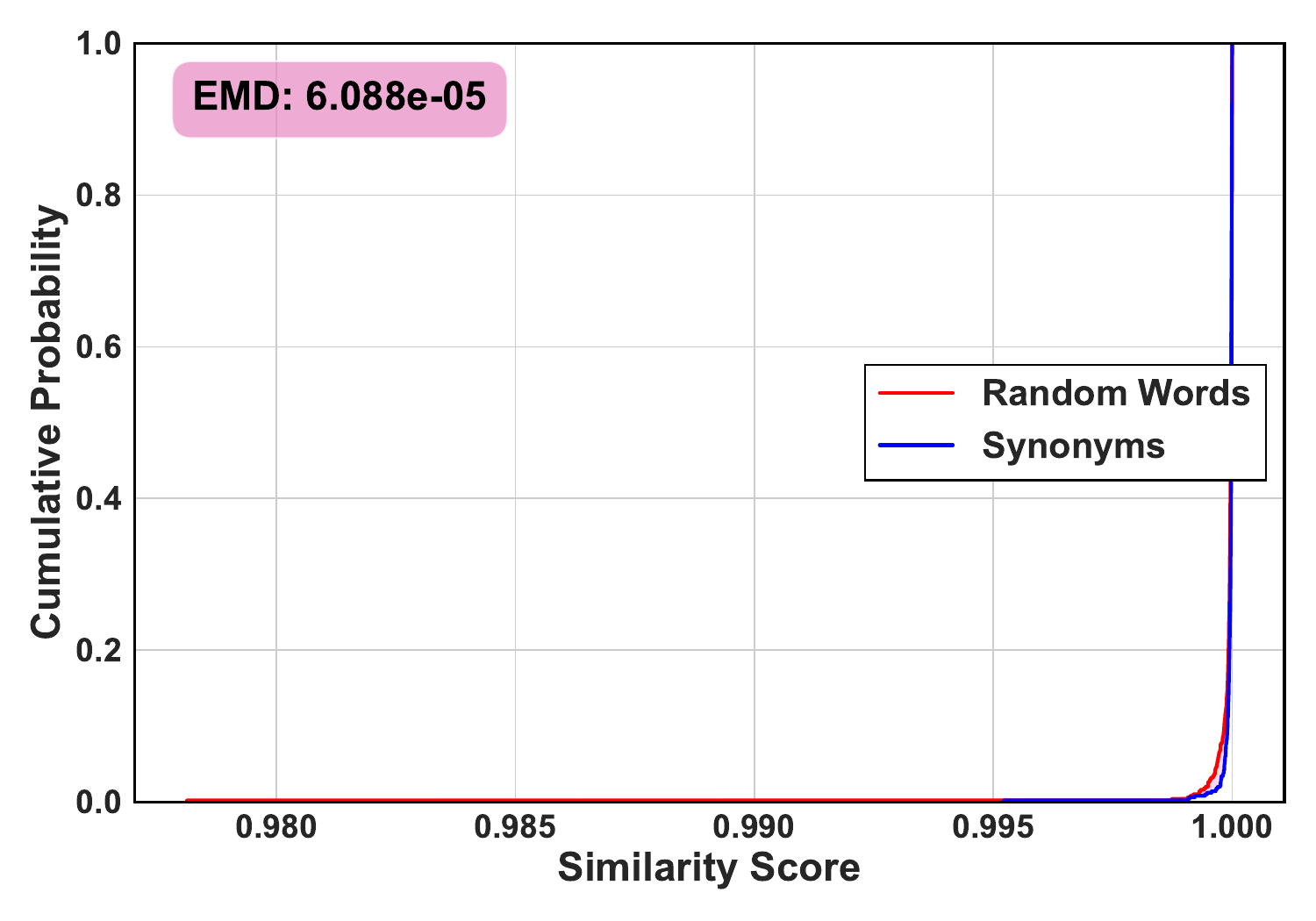}             
	\end{tabular}}
	&
	\subfloat[EOS and Cosine]
	{\label{fig:1b}%
		\begin{tabular}[b]{c}
			\includegraphics[width=0.31\textwidth]{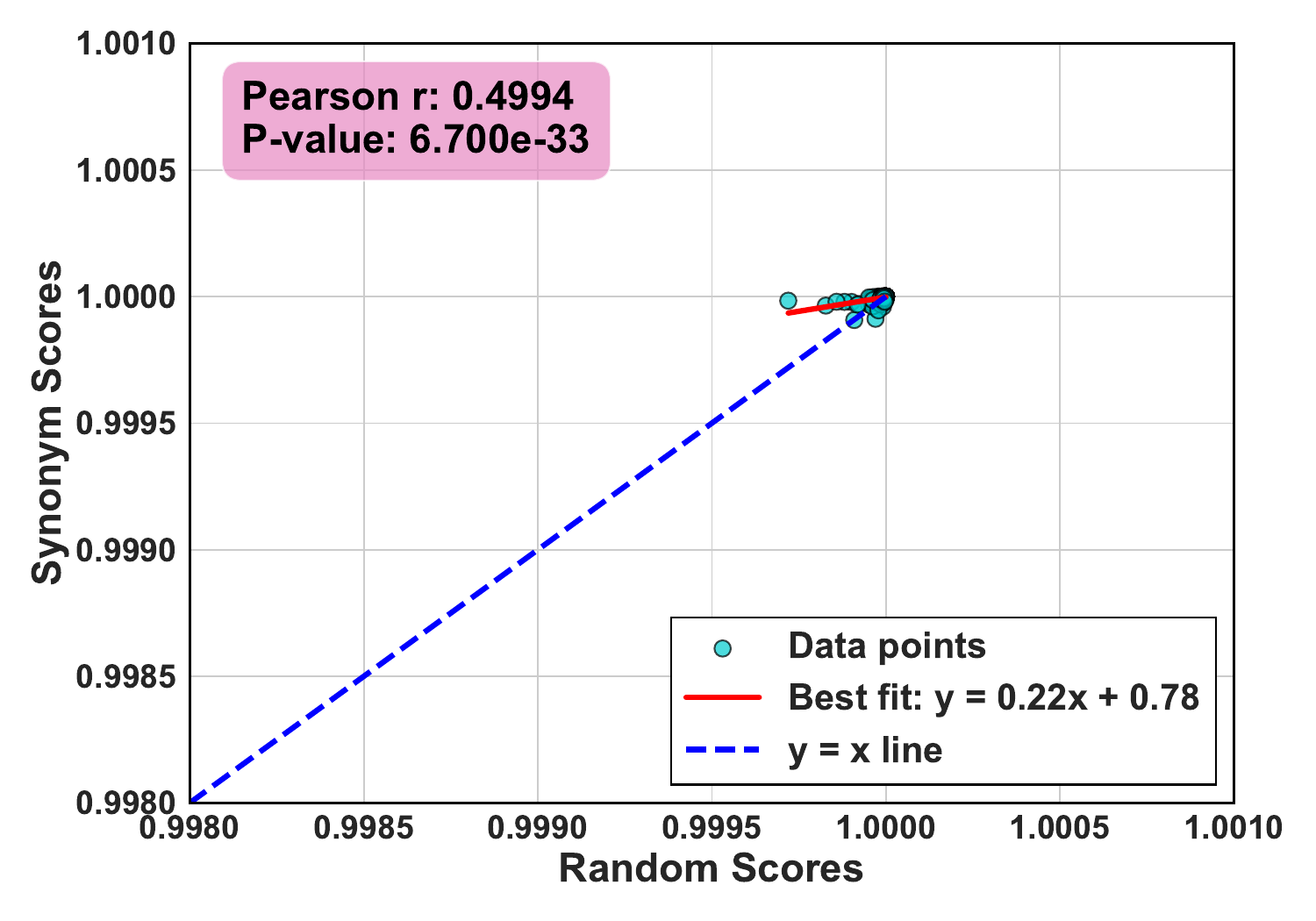} \\[2mm]
			\includegraphics[width=0.31\textwidth]{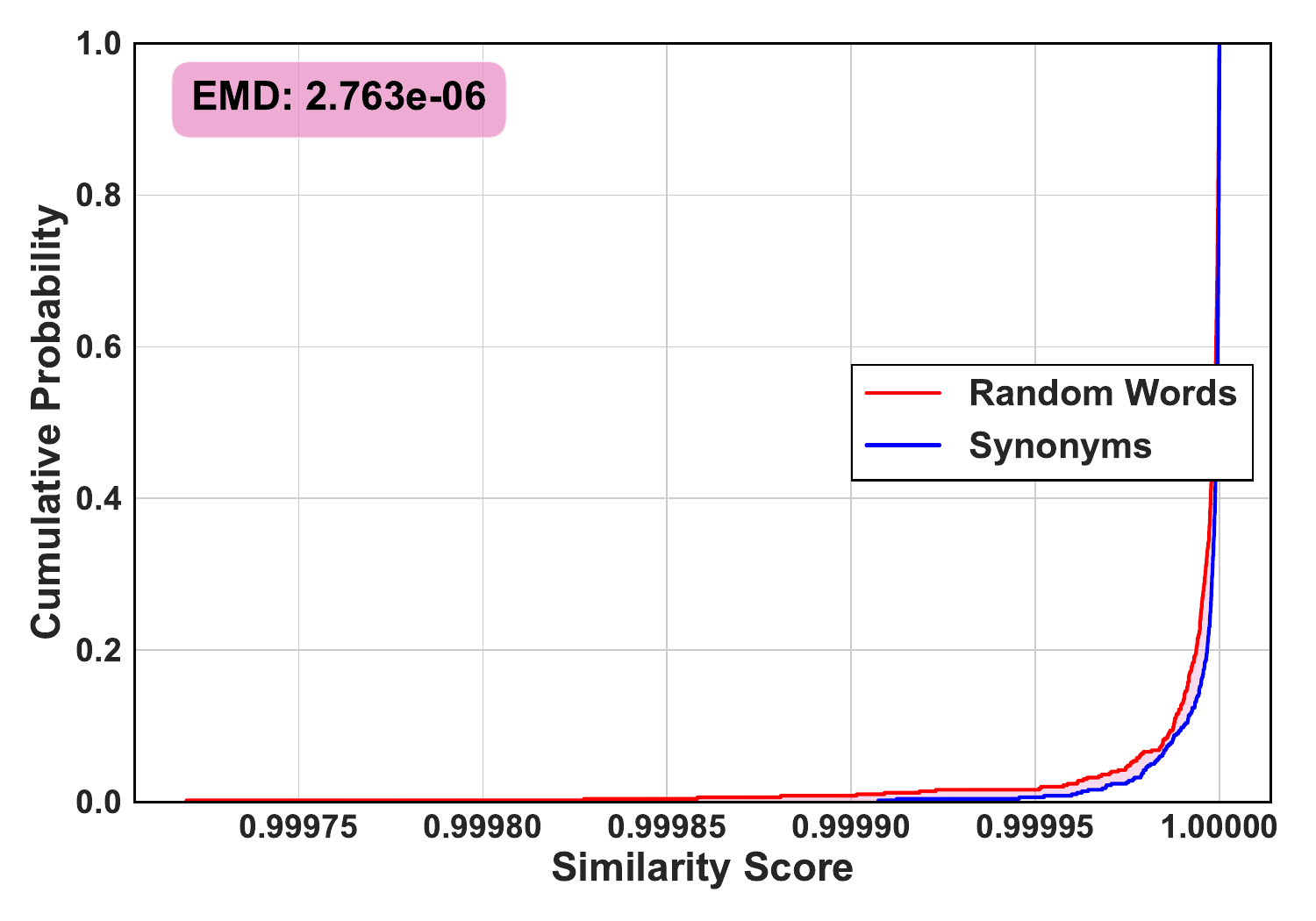}             
		\end{tabular}}
	&
	\subfloat[DDR]{\label{fig:1c}
		\begin{tabular}[b]{c}
			\includegraphics[width=0.31\textwidth]{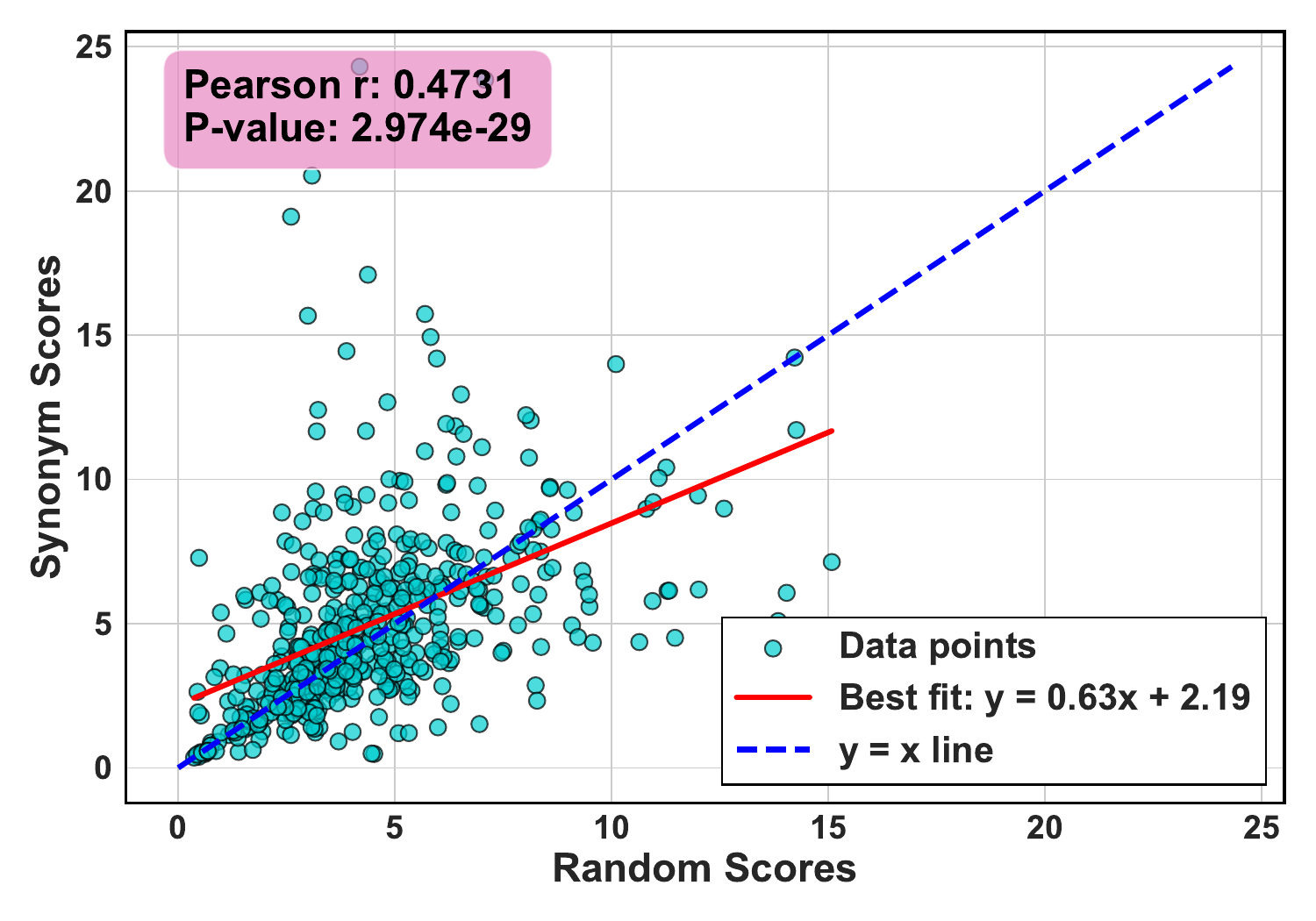} \\[2mm]
			\includegraphics[width=0.31\textwidth]{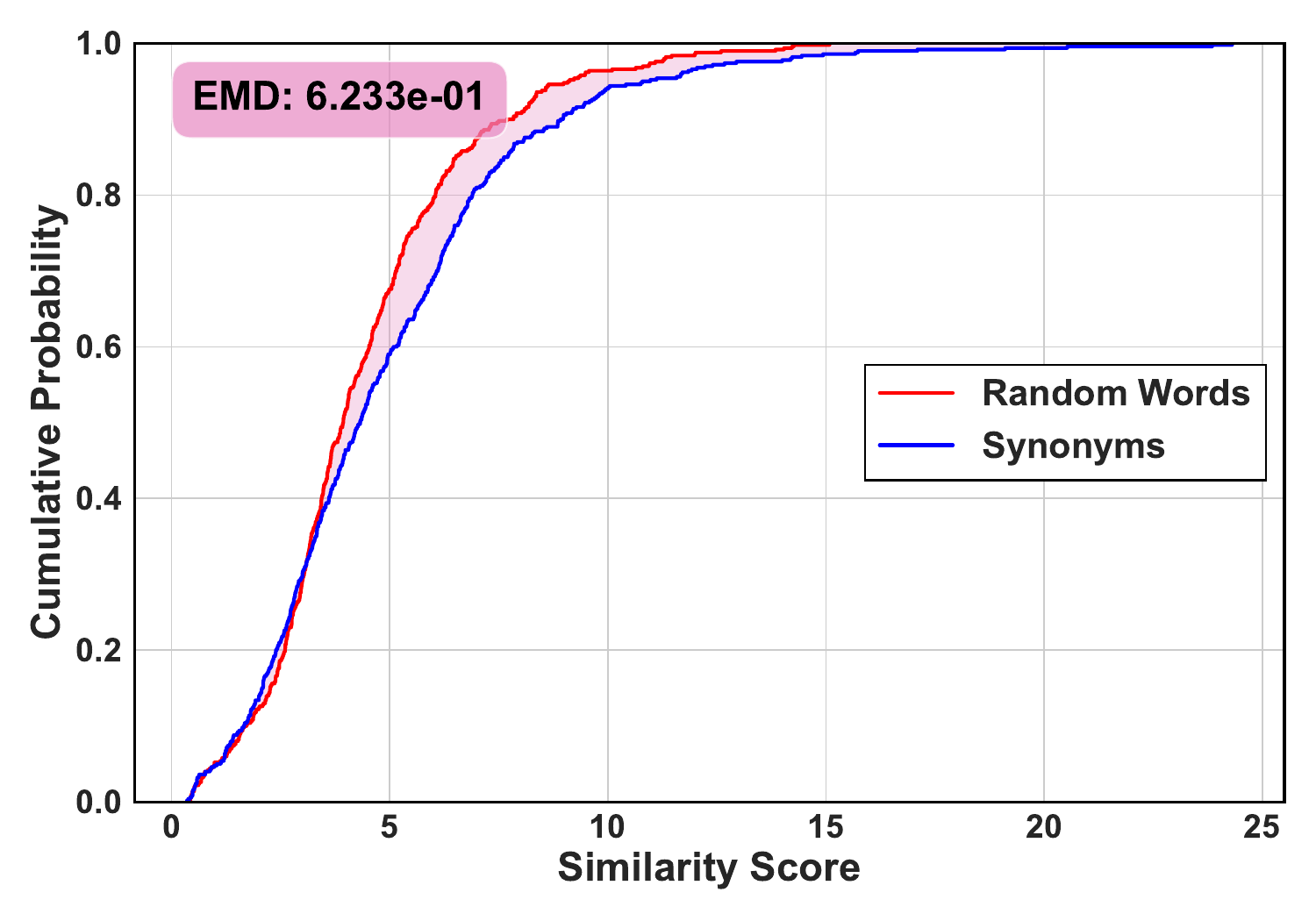}      
		\end{tabular}}
	\end{tabular}
	\caption{Comparison between methods on one edit}
	\label{fig:one_edit}
\end{figure}
\begin{figure}[!ht]
	\centering
	\begin{tabular}[c]{ccc}
		\subfloat[Centroid and Cosine]{\label{fig:2a}                                       
		\begin{tabular}[b]{c}                                                               
		\includegraphics[width=0.31\textwidth]{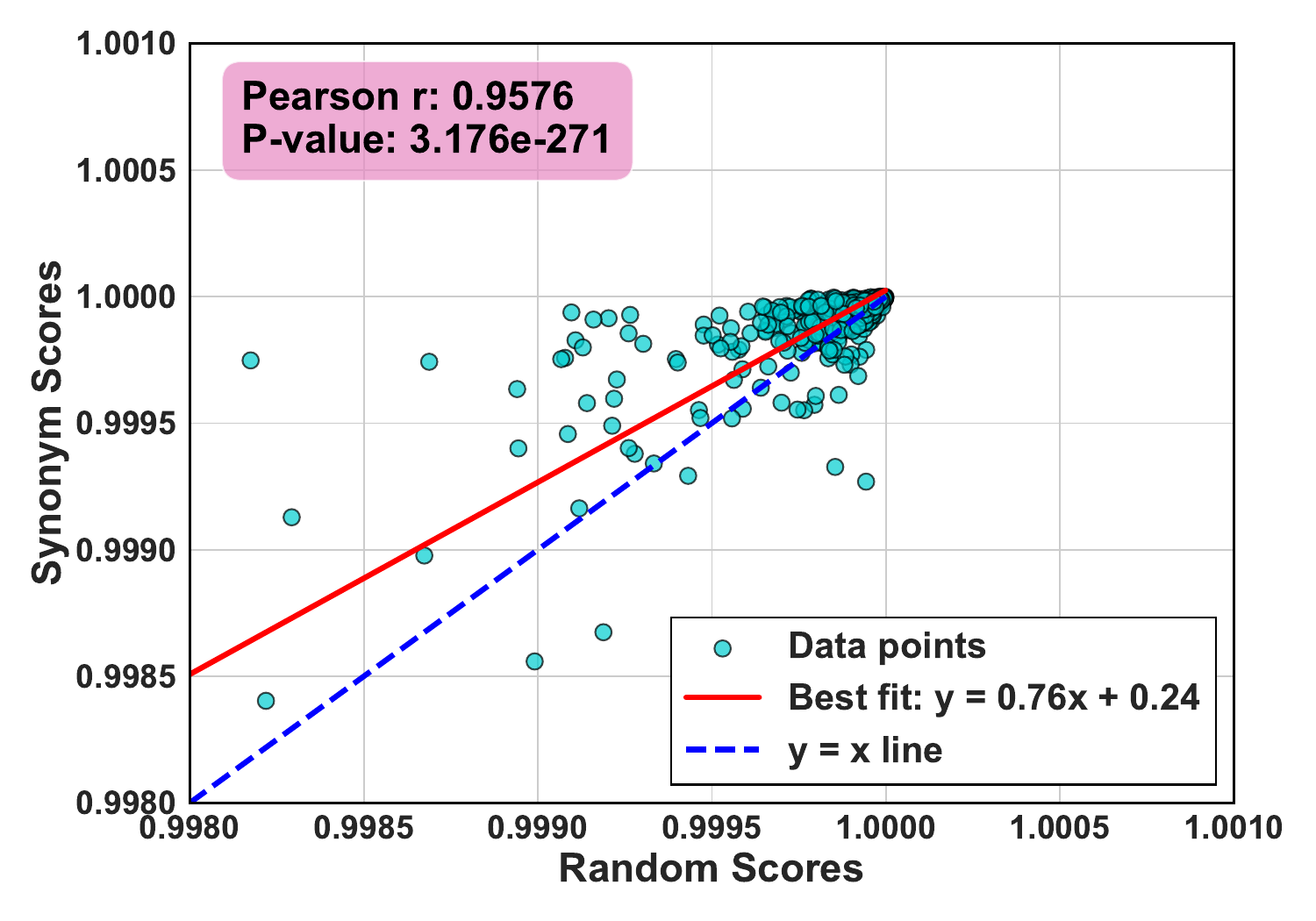} \\[2mm]
		\includegraphics[width=0.31\textwidth]{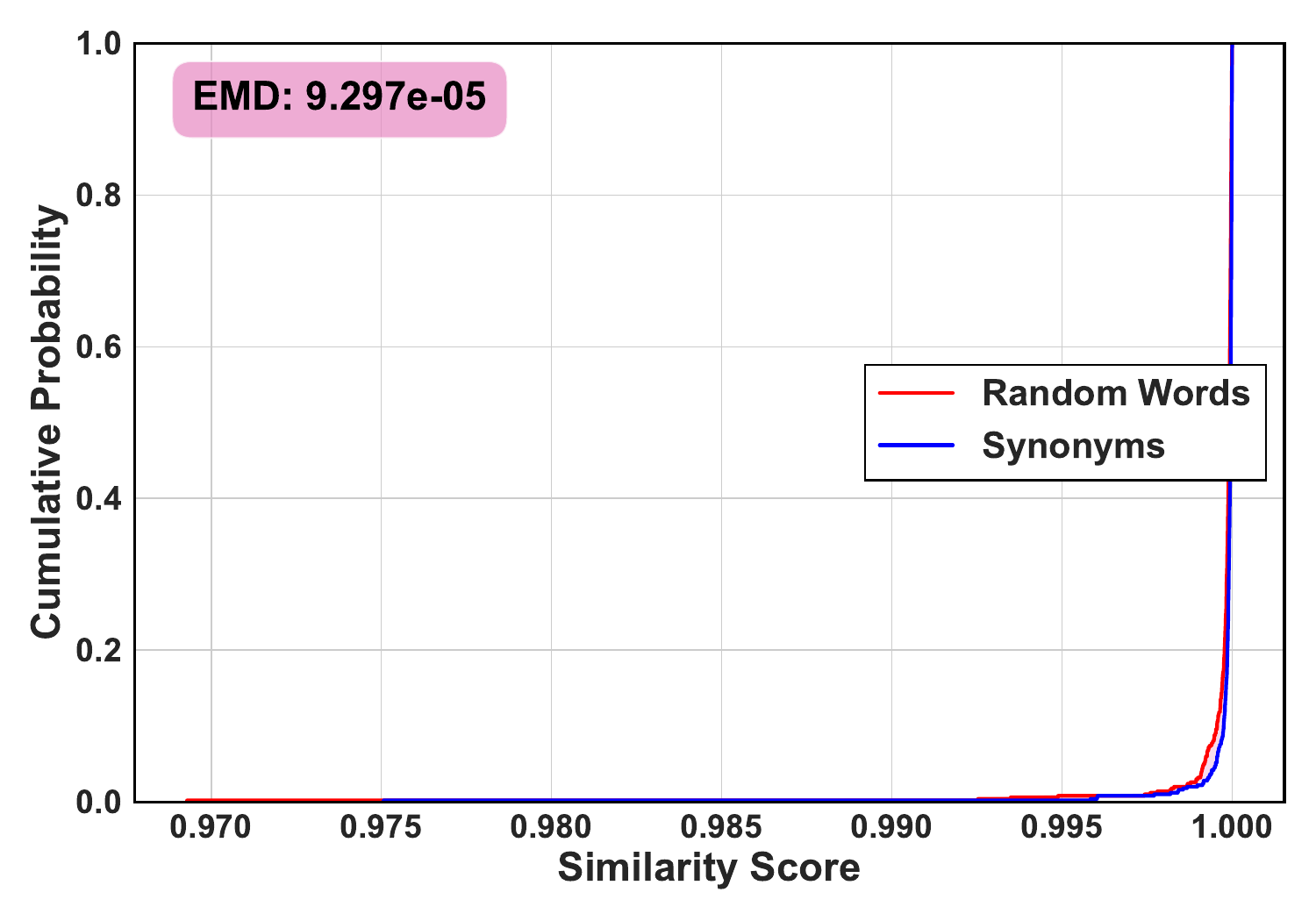}             
	\end{tabular}}
	&
	\subfloat[EOS and Cosine]
	{\label{fig:2b}
		\begin{tabular}[b]{c}
			\includegraphics[width=0.31\textwidth]{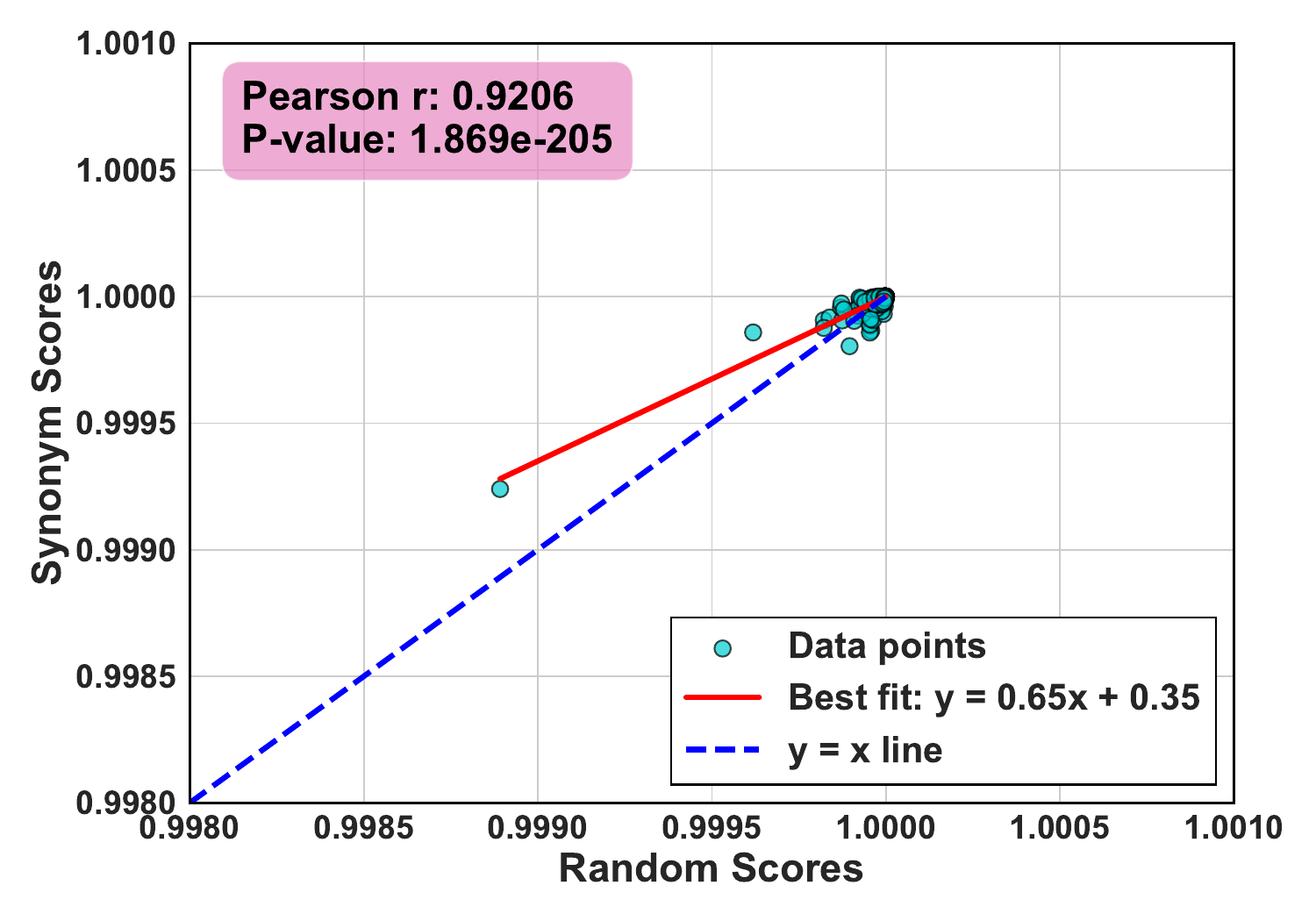} \\[2mm]
			\includegraphics[width=0.31\textwidth]{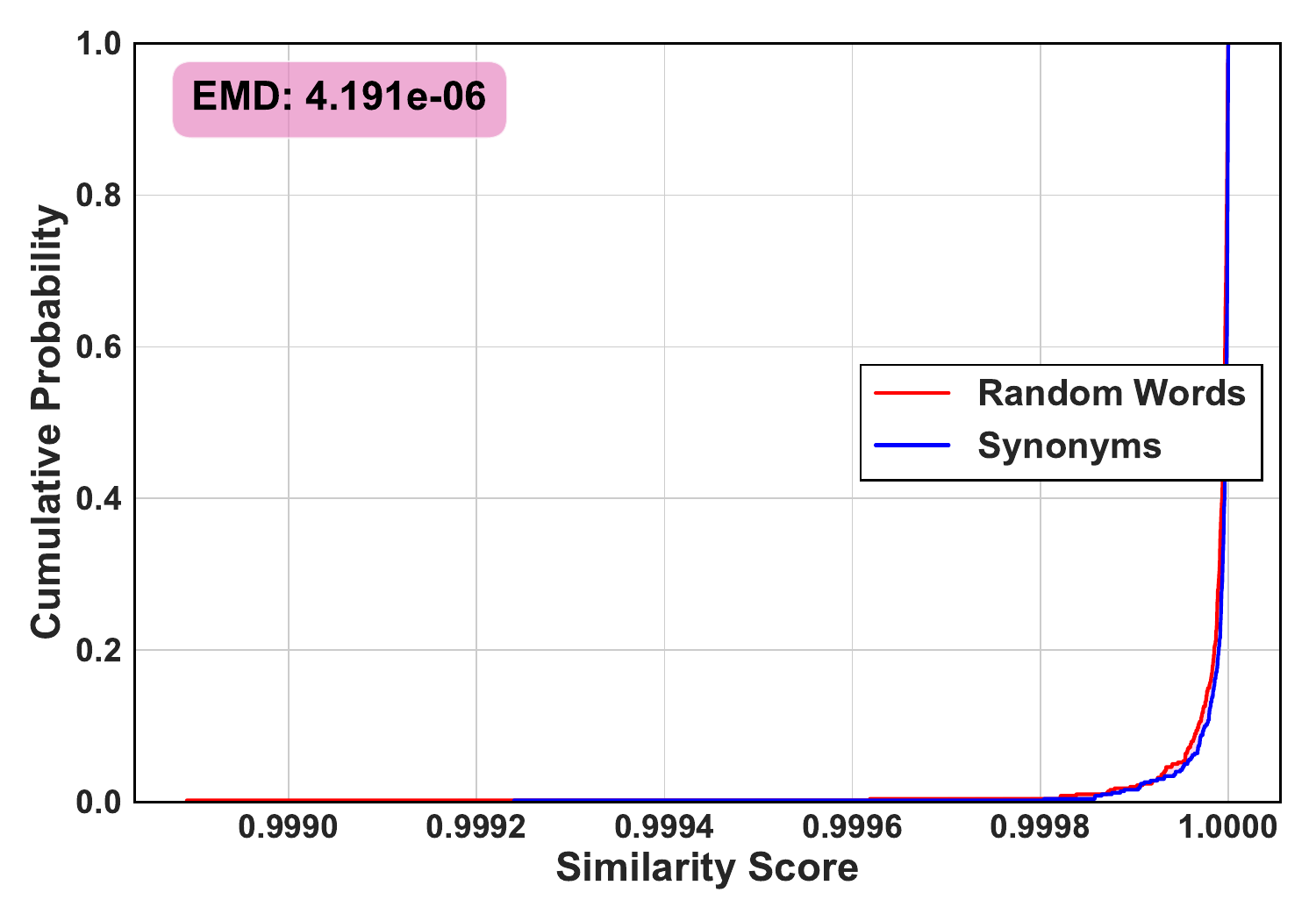}             
		\end{tabular}}
	&
	\subfloat[DDR]{\label{fig:2c}
		\begin{tabular}[b]{c}
			\includegraphics[width=0.31\textwidth]{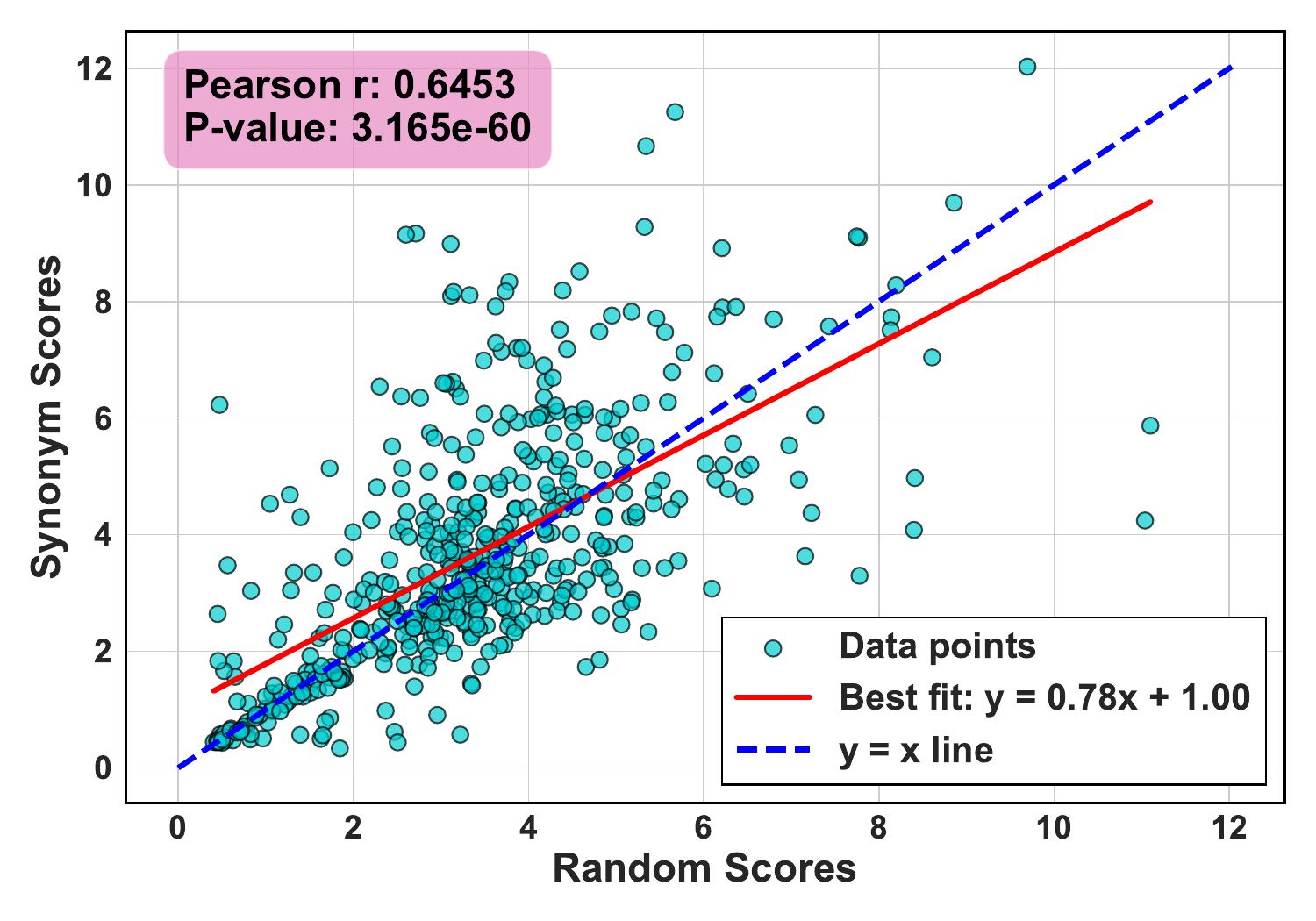} \\[2mm]
			\includegraphics[width=0.31\textwidth]{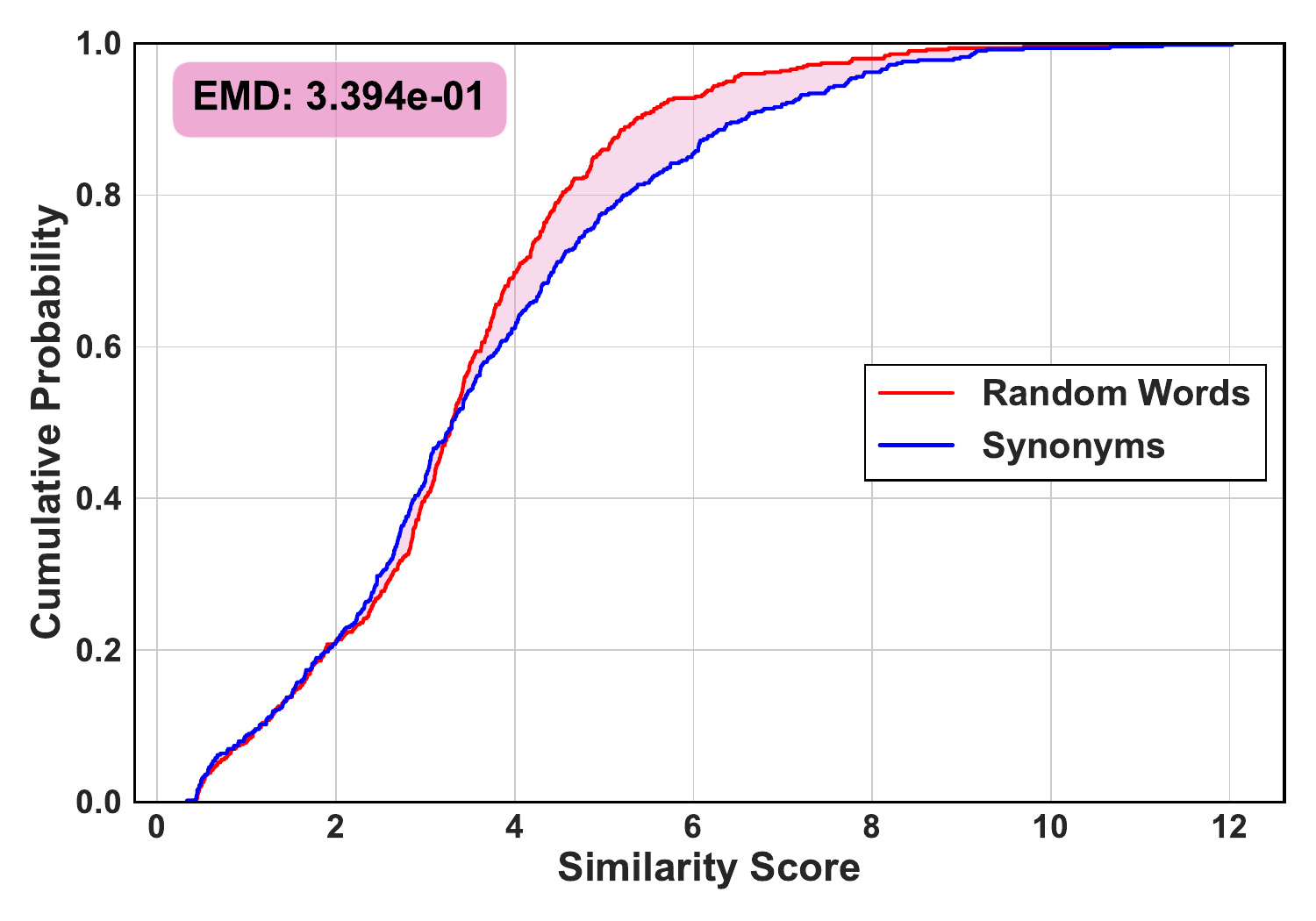}      
		\end{tabular}}
	\end{tabular}
	\caption{Comparison between methods on two edits}
	\label{fig:two_edits}
\end{figure}
\begin{figure}[!ht]
	\centering
	\begin{tabular}[c]{ccc}
		\subfloat[Centroid and Cosine]{\label{fig:3a}                                       
		\begin{tabular}[b]{c}                                                               
		\includegraphics[width=0.31\textwidth]{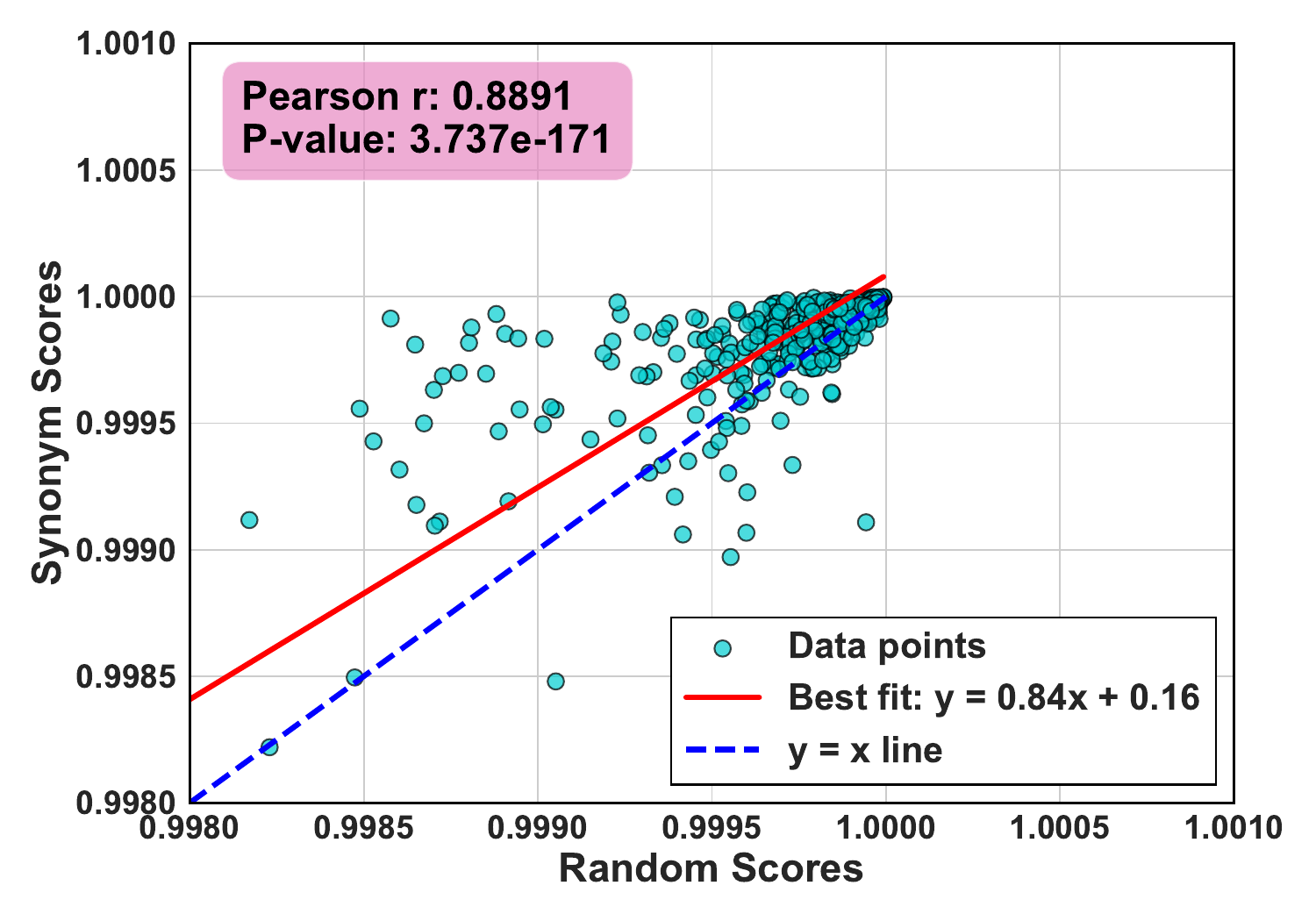} \\[2mm]
		\includegraphics[width=0.31\textwidth]{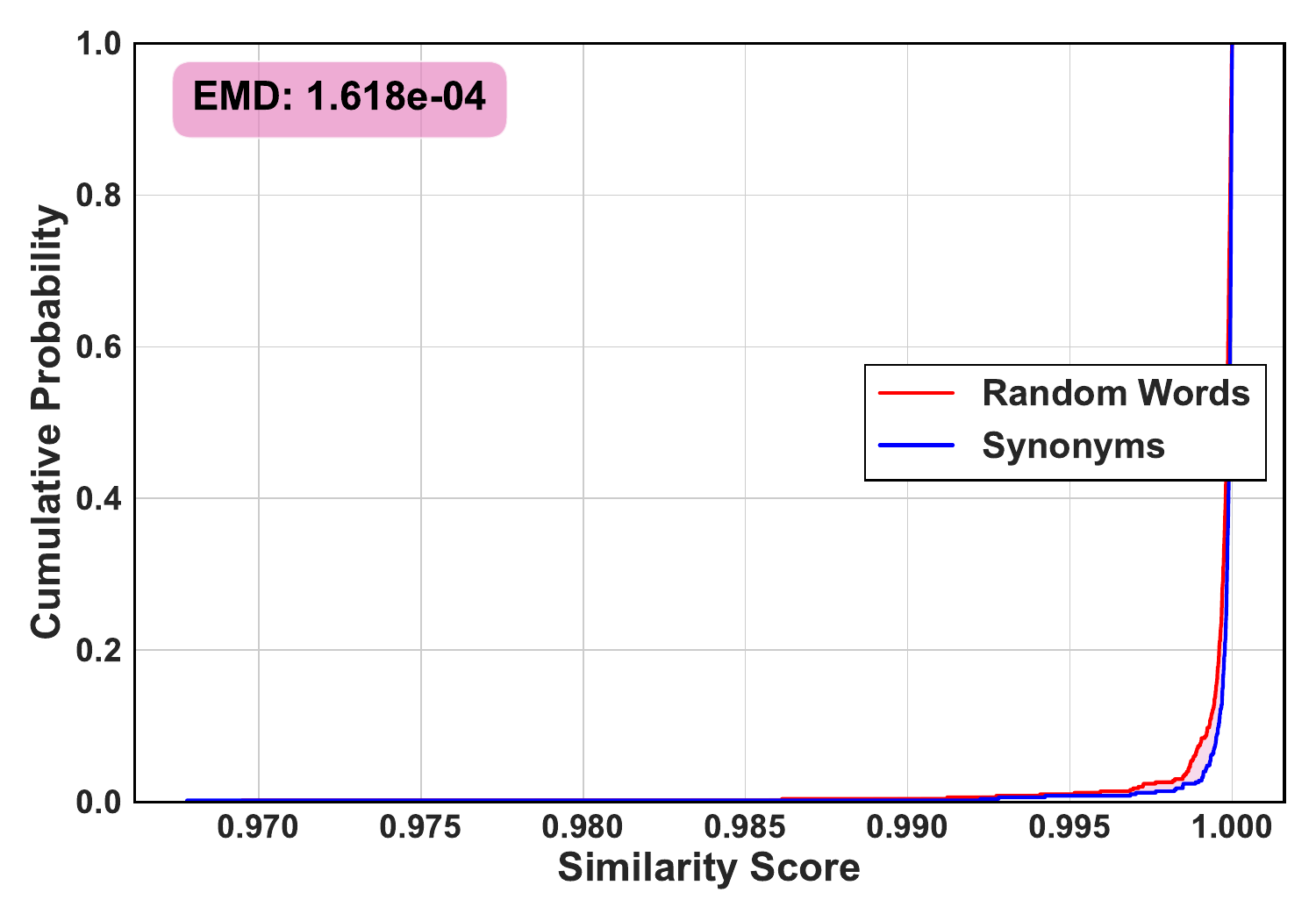}             
	\end{tabular}}
	&
	\subfloat[EOS and Cosine]
	{\label{fig:3b}
		\begin{tabular}[b]{c}
			\includegraphics[width=0.31\textwidth]{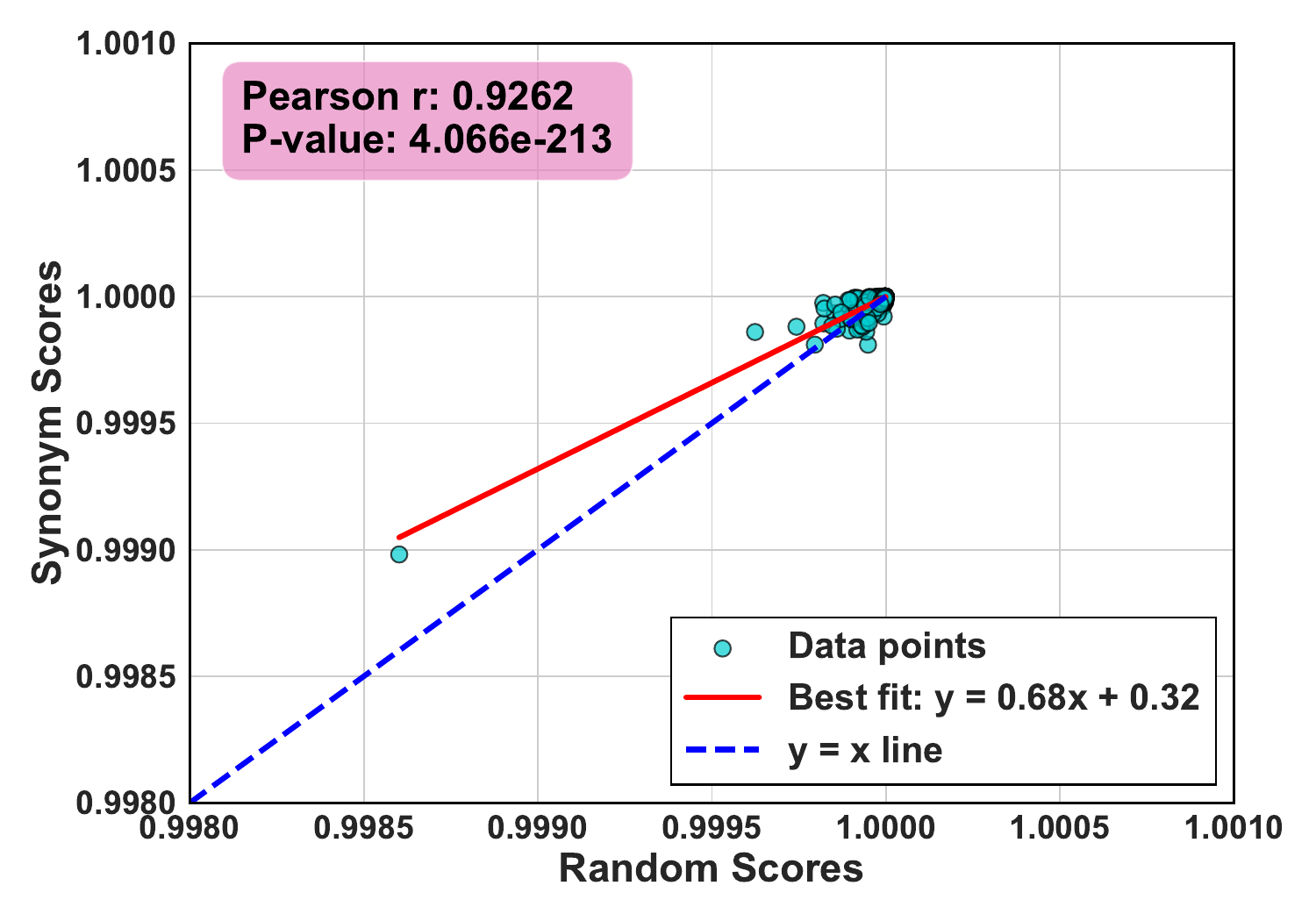} \\[2mm]
			\includegraphics[width=0.31\textwidth]{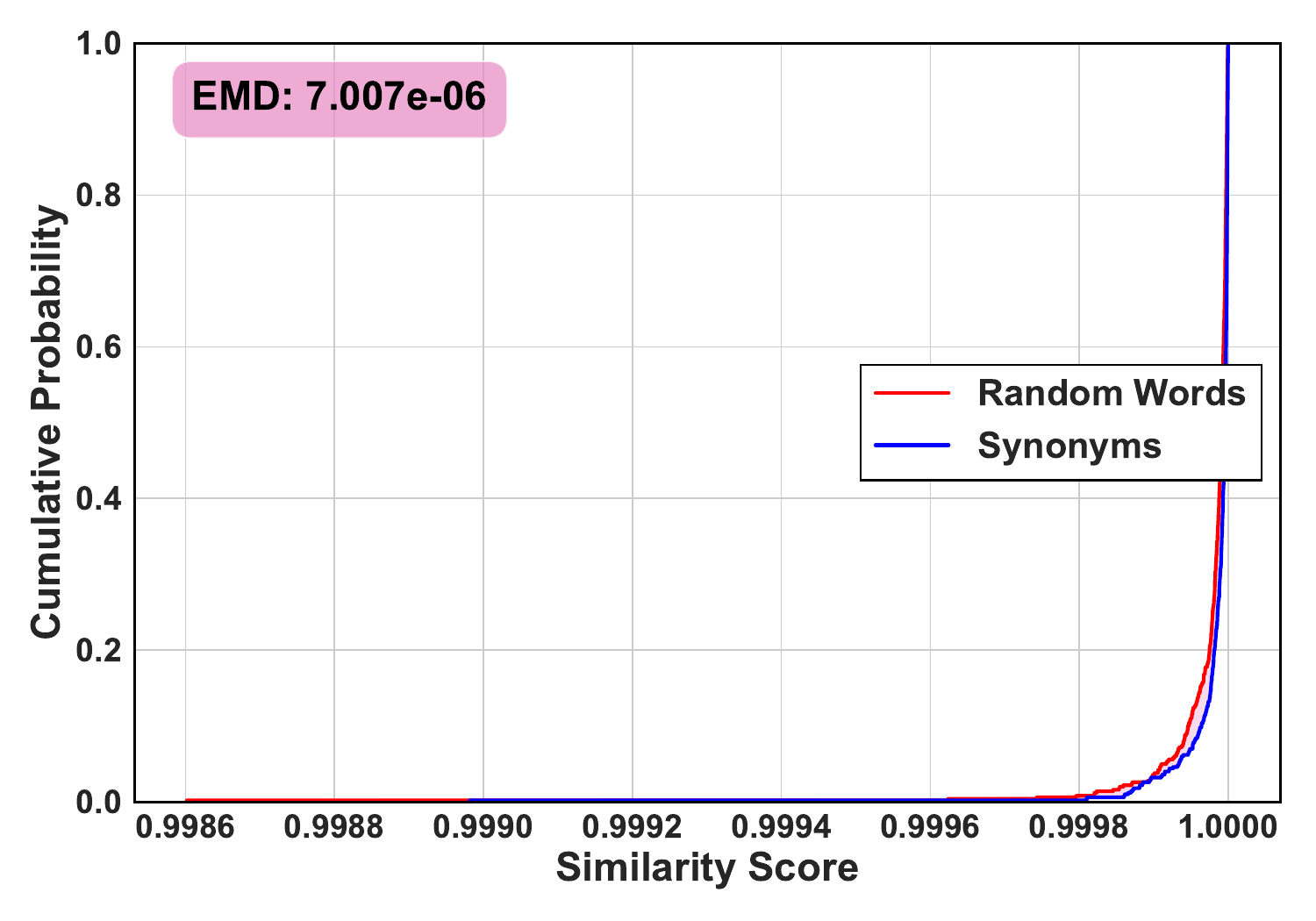}             
		\end{tabular}}
	&
	\subfloat[DDR]{\label{fig:3c}
		\begin{tabular}[b]{c}
			\includegraphics[width=0.31\textwidth]{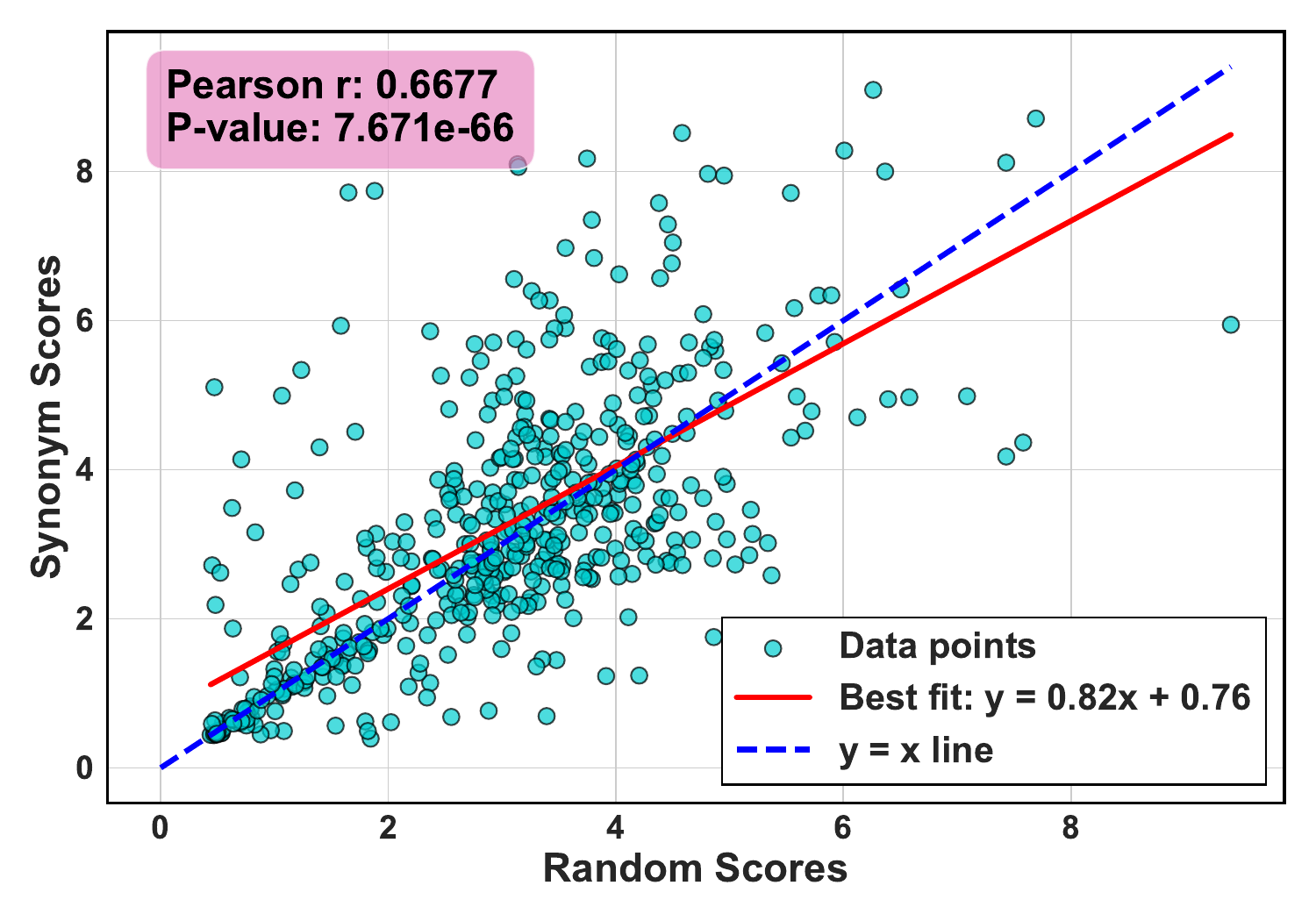} \\[2mm]
			\includegraphics[width=0.31\textwidth]{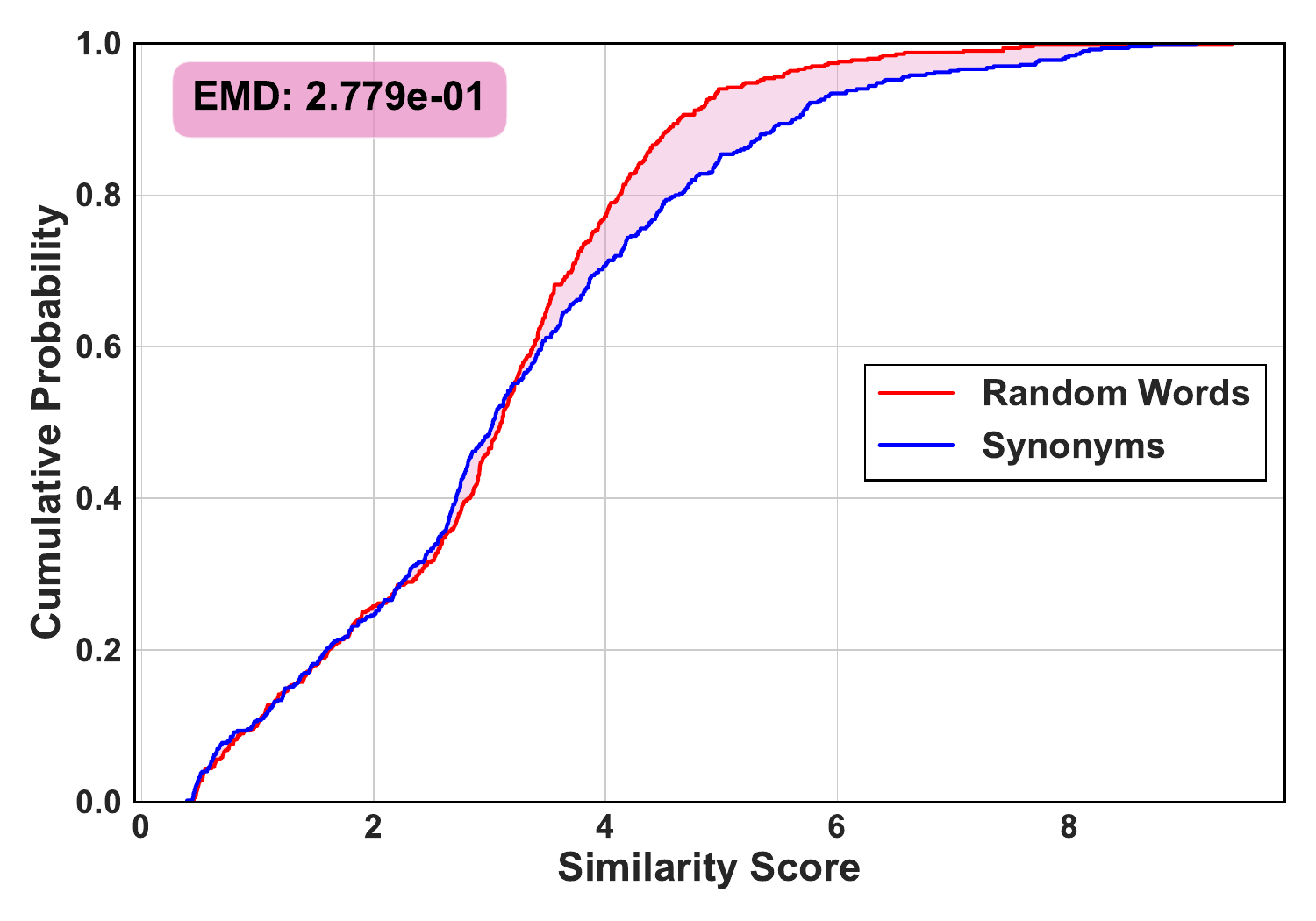}      
		\end{tabular}}
	\end{tabular}
	\caption{Comparison between methods on three edits}
	\label{fig:three_edits}
\end{figure}

The Centroid method (Fig.~\ref{fig:1a}) exhibits minimal differentiation, shown by points clustering closely around the diagonal line (Pearson $r \approx 0.9424$). Overlapping CDFs and negligible EMD ($\approx \num{6.088e-5}$) indicates almost no sensitivity to local semantic changes. When looking at two (Fig.~\ref{fig:2a}) and three (Fig.~\ref{fig:3a}) edits we see a similar pattern of minimal differentiation.

The EOS method (Fig.~\ref{fig:1b}) also shows minimal sensitivity. The EMD score ($ \approx \num{2.763E-6}$) is considerably smaller than that of Centroid, points deviate from the diagonal line with a reduced correlation ($r \approx 0.4994$), and the synonym-substitution distribution shifts slightly toward higher similarity (lower distance). The moderate Pearson score suggests that EOS embeddings partially capture semantic anomalies from random substitutions; however, the overlapping CDFs and lower EMD indicate the separation between distributions is minimal. Furthermore, EOS correlation quickly increases to $r \approx 0.9206 $ at two edits (Fig.~\ref{fig:2b}) and $ r \approx 0.9262$ at three edits (Fig.~\ref{fig:3b}), indicating the EOS method shows an abrupt decrease in sensitivity as edits accumulate.

The DDR method provides a clear semantic distinction at one-word edits (Fig.~\ref{fig:1c}). Scatter plots exhibit pronounced dispersion (Pearson $r = 0.4731$), and the synonym-substitution distribution shifts substantially compared to random substitutions, resulting in a much higher EMD ($\approx \num{6.233E-1}$). Unlike EOS and Centroid, DDR consistently maintains strong discriminative capability across edit depths. At two (Fig.~\ref{fig:2c}) and three (Fig.~\ref{fig:3c}) edits, DDR correlation moderately rises (to $r \approx 0.6453$ and $ \approx 0.6677$ respectively), but remains well below EOS and Centroid, while maintaining clearly separated distributions and large EMD values ($\approx \num{3.394E-1}$ at two edits and $\approx \num{2.779E-1}$ at three). Thus, DDR provides consistent, smooth, and meaningful differentiation as semantic edits accumulate, whereas Centroid remains insensitive and EOS rapidly loses discriminative power.

\section{Conclusion}
\label{sec:conclusion}
Our results show that DDR is a viable and practical method for measuring semantic (dis)similarity between text embeddings and outperforms common methods that rely on context embedding alone. One limitation of the current implementation is the requirement that sequences have the same length. To broaden the applicability of DDR, future work should generalize the distance computation to handle variable-length inputs. Such extensions will expand DDR's theoretical reach to enable practical applications. For example, DDR's sensitivity to small textual variations suggests potential uses in information retrieval (distinguishing near-duplicate queries) and evaluating model robustness to adversarial edits (detecting if small lexical edits flip the model's response when it shouldn't). Overall, DDR allows for an expanded range; that is, similarity scores occupy a broader interval and discrimination at finer resolution.

\end{document}